\documentclass{ieeetmlcn}
\usepackage{amsmath,amssymb,amsfonts}
\usepackage{algorithmic}
\usepackage{graphicx,color}
\usepackage{textcomp}
\usepackage{xcolor}
\usepackage{hyperref}
\usepackage{diagbox}
\usepackage{tikz}
\usetikzlibrary{chains,scopes,positioning,backgrounds,shapes,fit,shadows,calc,arrows.meta}
\usepackage{verbatim}
\usepackage{algorithm,algorithmic}
\def\BibTeX{{\rm B\kern-.05em{\sc i\kern-.025em b}\kern-.08em
    T\kern-.1667em\lower.7ex\hbox{E}\kern-.125emX}}
\AtBeginDocument{\definecolor{tmlcncolor}{cmyk}{0.93,0.59,0.15,0.02}\definecolor{NavyBlue}{RGB}{0,86,125}}

\def\OJlogo{\vspace{-4pt}\includegraphics[height=18pt]{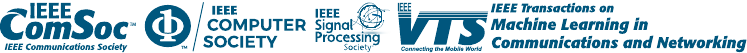}}
\def\seclogo{\vspace{10pt}\includegraphics[height=18pt]{new_logo}}

\def\authorrefmark#1{\ensuremath{^{\textbf{#1}}}}

\begin{document}
\receiveddate{XX Month, XXXX}
\reviseddate{XX Month, XXXX}
\accepteddate{XX Month, XXXX}
\publisheddate{XX Month, XXXX}
\currentdate{XX Month, XXXX}
\doiinfo{TMLCN.2022.1234567}

\markboth{}{Author {et al.}}

\title{Feature selection algorithm based on incremental mutual information and cockroach swarm optimization (February 2023)}

\author{First \authorrefmark{1},  Second Wei. Chen\authorrefmark{2} }
\affil{National Institute of Standards and Technology, Boulder, CO 80305 USA}
\affil{Department of Physics, Colorado State University, Fort Collins, CO 80523 USA}
\affil{Electrical Engineering Department, University of Colorado, Boulder, CO 80309 USA}
\corresp{Corresponding author: First A. Author (email: author@ boulder.nist.gov).}
\authornote{This paragraph of the first footnote will contain support information, including sponsor and financial support acknowledgment. For example, ``This work was supported in part by the U.S. Department of Commerce under Grant 123456.''}

\begin{abstract}

Feature selection is an effective preprocessing technique to reduce data dimension. For feature selection, rough set theory provides many measures, among which mutual information is one of the most important attribute measures. However, mutual information based importance measures are computationally expensive and inaccurate, especially in hypersample instances, and it is undoubtedly a NP-hard problem in high-dimensional hyperhigh-dimensional data sets. Although many representative group intelligent algorithm feature selection strategies have been proposed so far to improve the accuracy, there is still a bottleneck when using these feature selection algorithms to process high-dimensional large-scale data sets, which consumes a lot of performance and is easy to select weakly correlated and redundant features. In this study, we propose an incremental mutual information based improved swarm intelligent optimization method (IMIICSO), which uses rough set theory to calculate the importance of feature selection based on mutual information. This method extracts decision table reduction knowledge to guide group algorithm global search. By exploring the computation of mutual information of supersamples, we can not only discard the useless features to speed up the internal and external computation, but also effectively reduce the cardinality of the optimal feature subset by using IMIICSO method, so that the cardinality is minimized by comparison. The accuracy of feature subsets selected by the improved cockroach swarm algorithm based on incremental mutual information is better or almost the same as that of the original swarm intelligent optimization algorithm. Experiments using 10 datasets derived from UCI, including large scale and high dimensional datasets, confirmed the efficiency and effectiveness of the proposed algorithm

\end{abstract}



\maketitle

\section{INTRODUCTION}
$\quad$Nowadays, big data is growing explosively in all walks of life. A large amount of data brings data uncertainty and redundancy, which brings huge challenges to almost all algorithms in the field of mechanical learning and data mining, and often leads to poor results in learning tasks. Traditional data mining algorithms will show pressure when dealing with big data, because of the need for larger storage space and computing performance. Therefore, many scholars have proposed various feature selection algorithms to remove irrelevant and redundant features in the data set and obtain key feature subsets, which greatly improve the operation time and classification accuracy of subsequent data set processing. Therefore, an effective feature selection algorithm is very necessary to select the key information features from the original data.

Rough set is a powerful tool for processing uncertain data. It was proposed by Z.Prawlak in 1982 [1] and is widely used in the fields of data mining, data processing, pattern recognition, etc. Its main idea is to use knowledge granularity information to divide indistinguishable individuals into multiple groups, and then divide rough information system into three regions according to equivalence classes. They are called positive region, negative region and boundary region respectively. Based on these three areas, classification induction, rule extraction and attribute reduction can be carried out. As an important research direction, many scholars have conducted a large number of studies from different perspectives, such as information entropy-based attribute reduction algorithm [4]-[5], differential matrix algorithm [12]-[13], mutual information based attribute reduction algorithm [6]-[10], incremental algorithm [27], information based attribute reduction algorithm, etc. The basic idea of these algorithms is to eliminate irrelevant and redundant features and retain key features by applying rough set method and various feature evaluation functions. These methods only need the information of the data itself, and do not need any other knowledge, so the rule extraction and classification induction based on rough sets are objective.

Among these application algorithms, as an effective tool for processing discrete data, the attribute reduction algorithm based on mutual information is widely used. Its basic idea is to redefine attribute importance from the perspective of information theory, and take it as heuristic information to reduce the search space of attribute reduction process. Various improved algorithms have been proposed by scholars. Liu. [7] proposed a sampling method based on image region information entropy, and in this method, the hybrid optimization algorithm combining Powell and simulated annealing was used to optimize the mutual information function, which improved the accuracy and speed of mutual information calculation. Xu, Miao, Wei et al.[8] Combined the membership function of fuzzy set and the knowledge entropy of rough set, applied it to the fuzzy environment, generalized the metric concept of mutual information, and made it able to evaluate the importance of attributes in the fuzzy decision table. In order to reduce the bias of mutual information to multi-valued attributes and limit its value within the interval, Estevez P A, Tesmer M, Perez C A, et al.[10] used average normalized mutual information as a measure of redundancy between features to filter features and select the optimal feature subset. However, the dependency relationship between feature groups cannot be found, and the weakly correlated but non-redundant features can be filtered out. In order to solve this problem, Sun, Song, Liu et al.[11] proposed a two-stage feature selection method based on feature sorting and approximate Markov blanket to analyze the correlation and redundancy of features respectively and select effective feature subsets.

In high-dimensional data sets, it is difficult to find the optimal feature subset by exhaustive search. The solutions obtained by the traditional heuristic search algorithm of rough sets often fall into the local optimal solution. Therefore, in order to obtain the global optimal solution, many scholars turn to the study of Swarm Intelligence Optimization Algorithm, which is a common algorithm of computational intelligence and has obvious advantages in solving NP-hard problems. Its basic theory is to imitate the behavior of animal groups and use the information exchange and cooperation between groups to achieve the purpose of optimization. Among them, the most representative swarm intelligent algorithms are ant colony optimization algorithm proposed by Marco Dorigo[59], which continuously optimizes until it finds the optimal result by simulating pheromone updating in the foraging path of ants. The artificial bee colony algorithm was proposed by Karaboga[60], which uses three kinds of bees corresponding to different honey collecting mechanisms to search for the optimal nectar source to achieve the optimal solution. Raj B, Ahmedy I, Idris M Y I, et al.[61] Combined bee colonies with genetic algorithm to achieve a balance between exploration and exploitation capabilities. Then particle swarm optimization algorithm [65] updates itself by tracking individual optimal extremum and global optimal extremum, so as to find the optimal solution. Song X F, Zhang Y, Gong D W, et al.[66] combined the advantages of correlation, clustering and particle swarm and applied them to feature selection, a new three-phase hybrid FS algorithm was proposed. Feng, J. and Z. Gong[56] combined neighborhood rough sets with particle swarm to improve the problem that traditional algorithms are prone to fall into local extremes.

The principle behind the integrated cockroach swarm algorithm in this paper is similar to that of the particle swarm optimization algorithm. Cheng L, Han L, Zeng X, et al.[70] improved the cockroach algorithm and applied it to the path planning problem of robots by updating its own position through tracking individual optimal and global optimal. Wu, S. J. and C. T.[71] Combined with genetic algorithm, applied to the computational optimization of S-type biological systems.

In this paper, we propose a model combining rough sets with cockroach swarm optimization algorithm and apply it to feature selection. First, we discuss the properties of rough sets for feature selection. Second, we discuss the feature evaluation and feature search strategies for feature selection. In attribute reduction, we reduce the incoming instance attribute set based on mutual information to reduce the attribute search space. This analysis optimizes random searches for cockroach populations to reduce unnecessary searches. Finally, a feature selection algorithm is designed for the feature selection problem. kNN classifier is used as the fitness value of updating individual optimal extremum and global optimal extremum of random algorithm and the standard of evaluating feature subset. We have listed the experimental results, which show that the algorithm proposed by us has less cardinality of the selected feature subset and less operation time, especially on high-dimensional data sets, under the condition that the classification accuracy is good, which proves that the algorithm is effective.

The main contributions of this study are as follows:

1) Firstly, the application of cockroach swarm algorithm to feature selection is studied, and a new feature selection model combining mutual information of rough sets with CSO is proposed.

2) The traditional CSO model has fast convergence speed and is easy to fall into local optimization. In order to enhance the global search ability of the optimization algorithm, we have made some improvements to the cockroach swarm algorithm of this model. Inspired by reverse learning, we used reverse learning to initialize the population. Secondly, when updating the population position, we combined tabu search to avoid repeated search, jump out of local optimal, and improve optimization efficiency.

3) The experiment was carried out on 10 different types of data, and the experimental results confirm that the proposed algorithm can greatly improve the computational running time and reduce the cardinality of feature subset under the condition that the classification accuracy of the data set is ensured.

This paper is organized as follows. In the second part, we introduce some basic concepts about rough sets and application principles of cockroach swarm optimization algorithm. Then, the third part introduces the rough set model for attribute reduction and the feature selection algorithm combined with improved cockroach swarm optimization. In the fourth part, we introduce the experimental results and analyze the effectiveness of the feature selection algorithm. Finally, the fifth section makes a conclusion to this paper.

\section{PRELIMINARIES}

This section introduces some basic knowledge of rough sets and cockroach swarm optimization algorithm for further discussion.

\subsection{ROUGH SETS}

In rough set theory, knowledge is described as an information system composed of row vectors and column vectors. The row vectors represent objects in the theoretical domain and the column vectors represent different attributes. One kind of information system is called decision table, which is composed of condition attribute and decision attribute, and each attribute has its corresponding attribute value. In a decision table, not every conditional attribute is necessary. Some conditional attributes are unrelated to the decision attribute, or are redundant, so they do not need to be added to the calculation. The attribute reduction theory of rough set is to deal with this problem. It removes unnecessary and redundant attributes while keeping the classification ability of information system unchanged.

Let the decision table be a non-empty finite set $(U,C \cup D)$, where $U={x_1,x_2,...,x_n}$ is called the domain, $C={a_1,a_2,...,a_n}$ is the conditional attribute set, and D is the decision attribute set. As shown in Table 1, the decision table consists of 4 samples, 3 conditional attributes {a1,a2,a3} and a decision attribute {d}.

\begin{table}
	\caption{An example of DS}
	\label{table}
	\setlength{\tabcolsep}{4pt}
	\begin{tabular}{p{40pt}p{40pt}p{40pt}p{40pt}p{40pt}}
		\hline   
		$U$& $a_1$& $a_2$& $a_3$& d \\
		\hline
		$ x_1 $ & 1& 0 & 1& 1 \\ 
		$ x_2 $ & 0& 1 & 0& 0 \\
		$ x_3 $& 1& 1 & 0& 1 \\
		$ x_4 $& 1& 0 & 1& 0 \\
		\hline
	\end{tabular}
\end{table}

The concepts commonly used in information theory include information entropy, conditional entropy, joint entropy, mutual information and so on. Before calculating mutual information, it is necessary to calculate the information entropy, conditional entropy or joint entropy of discrete random variable a. First, the definition of information entropy is given.

1) Shannon defines information entropy as the occurrence probability of discrete random events. Large probability, more chances, less uncertainty; On the contrary, the uncertainty is large. Let the information entropy be H(X), which is defined as follows:

\begin{equation}
	\label{eq1}
	 H(X) = - \sum_{x=X} P(x)  log_{2} P(x)
\end{equation}
In formula (1), $ P(x) = \frac{ |x_i| }{ |U| } , i = \{1,2,...,n\} $ , $|U|$ is the base of U, said the appear probability of discrete values of $x$ variables, assuming that variable $a_1 = [1,0,1,1] $ and $a_2=[0,1,1,0]$,
According to probability
$p(a_1=0) = \frac{1}{4};  p(a_1=1) = \frac{3}{4} ;
p(a_2=0) = \frac{2}{4};  p(a_2=1) = \frac{2}{4} $, so that,  
$H(a_1) = -\frac{1}{4} * log (\frac{1}{4} )  -\frac{3}{4} * log (\frac{3}{4} )   =  0.5623 .$  Similarly, $H(a_2) = 0.6931.$

2) The joint information entropy of two variables X and Y [5] is defined as:

\begin{equation}
	\label{eq1}
	H(X,Y) = - \sum_{x=X}  \sum_{y=Y} P(x,y)  log_{2} P(x,y)
\end{equation}
Where x and y are specific values of X and Y, and correspondingly, P(x,y) is the joint probability that these values occur together.
\begin{table}
	\caption{Examples:}
	\label{table}
	\setlength{\tabcolsep}{4pt}
	\begin{tabular}{p{40pt}p{40pt}p{40pt}p{40pt}p{40pt}}
		\hline   
		$a_1$&
		$1$&
		$0$&
		$1$&
		1 \\
		\hline
		$ a_2 $&
		0& 1 & 1& 0 \\
		\hline
	\end{tabular}
\end{table}
According to Table 2, $
P(a_1=0, a_2=0) = 0,
P(a_1=0, a_2=1) = \frac{1}{4},
P(a_1=1, a_2=0) = \frac{2}{4},
P(a_1=1, a_2=0) = \frac{1}{4},$
By calculating $H(a_1,a_2)$, we get:
$  H(a_1,a_2) =   -\frac{1}{4} log(\frac{1}{4} )  -\frac{2}{4} log(\frac{2}{4} )  -\frac{1}{4} log(\frac{1}{4} )  =  1.0397.  $

3). The conditional entropy H (Y | X) under the condition of known random variable X, the uncertainty of a random variable Y. It is defined as:

\begin{equation}
	\label{eq1}
	H(Y|X) = - \sum_{x=X} P(x) \sum_{y=Y} P(y|x) log_{2} P(y|x)
\end{equation}
According to the probability of $(a_1,a_2)$ :
\begin{table}
	\caption{Examples:}
	\label{table}
	\setlength{\tabcolsep}{4pt}
	\begin{tabular}{p{50pt}p{50pt}p{50pt}p{50pt}}
		\hline   
		\diagbox{$a_2$}{$a_1$} &
		$0$&
		$1$&
		$P(a_1 = i)$ \\
		\hline
		$ 0 $&
		0 & $\frac{1}{2}$ &  $\frac{1}{2}$ \\
		$ 1 $&
		$\frac{1}{4}$  & $\frac{1}{4}$  & $\frac{1}{2}$  \\
		$ P(a_2=i) $&
		$\frac{1}{4}$  & $\frac{3}{4}$  & 1 \\
		\hline
	\end{tabular}
\end{table}
$P(a_1 = 1 | a_2 = 1)$ said in the case of $a_2 = 1$ the probability of $a_1 = 1$, calculated as $   P(a_1=1|a_2=1) = \frac{P(a_1=1|a_2=1)}{P(a_2=1)}  = \frac{1/4}{2/4}  =  \frac{1}{2} ,$
similarly $P(a_1=0 | a_2=0), P(a_1 = 0 | a_2 = 1), P(a_1 = 1 | a_2 = 0)$.
The conditional entropy $H(a_1 | a_2)$ can be obtained.

4). Mutual information is a useful information measure, which can be regarded as the amount of information about random variable Y contained in random variable X [4]. Two random variables (X, Y) the joint distribution of $p(X, Y)$, the marginal distribution of $p(X)$, $p(Y)$ and mutual information $I(X, Y)$ is the joint distribution $p(X)p(Y)$ and marginal distribution $p(X)$, $p(Y)$ of the relative entropy [6], namely:
\begin{equation}
	\label{eq1}
	I(X,Y) = \sum_{x=X}  \sum_{y=Y}  P(x,y)  log_{2}  \frac{P(x,y)}{P(x)P(y)} 
\end{equation}
Common mutual information calculation formulas are as follows:
\begin{equation}
	\label{eq1}
	I(X;Y) = H(X) + H(Y) - H(X,Y)
\end{equation}

\subsection{COCKROACH SWARM OPTIMIZATION}

Inspired by the foraging behavior of cockroaches, the CSO algorithm is established by imitating the pursuit behavior of individual cockroaches to find the global optimal value. Its basic idea is to search for the optimal solution through continuous cooperation and information sharing between groups.

1) The basic idea of cockroach swarm algorithm

In the process of foraging, cockroaches will constantly look for the individual optimal cockroach and the global optimal cockroach, compare the advantages and disadvantages of the target solution, move towards the individual optimal cockroach or the global optimal cockroach, and constantly update their position and fitness value. If the fitness value is higher, they will replace the original individual. Keep track of the best locations you've found so far, and update your individual optimal location the next time you search for food. The optimal position of each cockroach in the crawling process is called the individual optimal pbest, and the optimal position of the whole population in the iterative process is called the global optimal gbest. The cockroaches share information through pbest and gbest, so as to change the search behavior of the population in the iterative process and achieve the optimal result. Cockroaches usually appear in dark and damp places, and their common foraging behaviors include chasing, gathering, scattering, cruelty, etc.

2) Update location

After randomly initializing a set of solutions, the CSO iteratively searches for the optimal solution, and the cockroaches constantly update their positions towards the two extreme values pbest and gbest through clustering behavior. The update rules are as follows:
\begin{equation}
	\label{eq1}
	y_r = \left\{
	\begin{aligned}
		y_r + a*rand*(\rho_r - y_r),  \quad  y_r \neq \rho_r \\
		y_r + a*rand*(\rho_g - y_r),  \quad  y_r = \rho_r 
	\end{aligned}
	\right.
\end{equation}
Where $y_r(r=1,2,3... ,n)$ is the rth cockroach position, $a$ represents the step size, is a fixed value, $rand$ is any value between $(0,1)$, $\rho_r   = opt_s \{ y_s,  | y_r  -   y_s |   \leq  visual  \}$ and $\rho_g   =  opt_r  \{  y_r    \}$ are the optimal position of the rth cockroach and the global optimal position, respectively. visual represents the visual field range of cockroaches, is a constant, $s=1,2,3... n$. After each round of foraging, randomly disperse each individual through the following model to maintain the current individual diversity, 

\begin{equation}
	\label{eq1}
	y_r   =   y_r   + rand(1, E ) ,   r = 1,2,...,N 
\end{equation}
$rand(1,E)$is an E-dimensional (problem space dimension) random vector that can be set within a certain range. Finally, Model (8) replaces randomly selected individuals with global optimality.

\begin{equation}
	\label{eq1}
	y_n =  \rho_g ,   n =  1,2,...,N
\end{equation}

\section{Feature selection model based on mutual information and improved CSO}

The basic idea of IMIICSO algorithm is to integrate mutual information, reverse learning, tabu search and CSO algorithm to search the best feature subset within a given number of iterations. Firstly, for the incoming original decision information system, the information relationship between the condition features and the decision features is calculated incrementally and stored in the matrix. Then, the initial population is constructed by reverse learning. Then, the improved CSO algorithm is used to search the optimal feature subset randomly, and kNN classifier is used as the subset evaluation standard. Some details of the model are described below.

\subsection{FEATURE REDUCTION}

The first step of the model is to use mutual information of rough set to preliminatively filter the feature set. If the mutual information size of feature a and decision feature d is 0 or negligible, a is directly discarded. In particular, for the characteristics of the two given a and d, features a information entropy H (a) and conditional information entropy H (a | d) difference is less than a threshold, so we decided that a condition attributes and decision attribute d relevance is very small, little effect on classification of decision. Therefore, these incoming conditional attributes do not need to be involved in feature selection, so that they can be filtered to select representative features. The process is shown in Algorithm 1:

\begin{algorithm}[H]
\caption{Feature reduction based on incremental mutual information of rough sets.}\label{alg:alg1}
\begin{algorithmic}[1]
\REQUIRE Decision table $DT = (U,C \cup D )$
\ENSURE  $DT_0 = (U,C_0 \cup D )$
\STATE split data

\WHILE{1 to len(data)}
\FOR{each $a \in C$}
\IF{$I(a;D) > threshold $}
\STATE $MI(i)  =  a $  
\ENDIF
\ENDFOR
\ENDWHILE

\STATE sort MI
\STATE $S_k$ = Filter the first partial feature subset according to the adjustable threshold
\STATE $C_0 = Sk $
\RETURN $DT_0$
\end{algorithmic}
\label{alg1}
\end{algorithm}

\subsection{SOLUTION EXPRESSION AND POPULATION INITIALIZATION COMBINED WITH REVERSE LEARNING}

Assume that the population size is $N$and the feature dimension is $D$. A solution is represented by a D-bit binary string, which corresponds to the feature set. 1 means that the feature is selected, and 0 means that the feature is not selected. Thus, a subset of iterative features is obtained. For example, a feature set $\{a, b, c \} $, if ultimately solution for $\{1, 1 \} $, the corresponding feature subset for $\{a, c \} $. Population initialization, randomly generated N * D D matrix X, the feasible solution of each solution is $X_g = \{x_ {g1}, x_ {g2}... x_{gn}\}, x_{gi} \in [0, 1]$;  The fitness of each solution is calculated and combined with the reverse learning idea proposed by Tizhoosh: in the worst case, the optimal solution obtained in the iterative process of the algorithm may be in the opposite position of the current position in the search space. Generate the corresponding inverse solution $X_g^* = \{x_{g1}^*,x_{g2}^*,... according to formula (9). x_{gn}^*\}$,
\begin{equation}
	\label{eq1}
	x_{gi}^* = \left\{
	\begin{aligned}
		1 - x_{gi},   \quad fitness(g) < mean \\
		x_{gi},    else
	\end{aligned}
	\right.
\end{equation}
Finally, the reverse solution $X_g^*$is evaluated individually, and N optimal solutions are selected from all solutions to form the initial population.

\subsection{Neighborhood search strategy based on pbest and gbest}

In the algorithm iteration, each iteration of cockroach i will constantly update its position according to the iterative optimal solution pbest and the global optimal solution gbest. The clustering behavior formula of cockroach swarm algorithm is improved and applied to the algorithm model. First, a solution was randomly selected and judged according to the fitness $f_i$of the ith cockroach. If the fitness is small, a local search is performed relative to pbest, and if the fitness is large, a new solution is generated relative to gbest global search. The formula is as follows:
\begin{equation}
	\label{eq1}
	y_i = \left\{
	\begin{aligned}
		y_i + a*rand*(\rho_i - y_k),   f_i \leq t \\
		y_i + a*rand*(\rho_g - y_k),   f_i > t 
	\end{aligned}
	\right.
\end{equation}

in Formula (10), $\rho_i$is the optimal solution of the current individual of the ith cockroach, $t \in (0,1)$is the fitness threshold, and $y_k, k \in [1, D] $is the KTH randomly selected solution.

\subsection{TABU SEARCH}

Tabu search is a global optimization algorithm first proposed by Glover, which simulates the process of human intelligence and memory. Tabu search algorithm uses some mechanisms to avoid repeated search so as to ensure the implementation of global solution construction. In IMIICSO, taboo search is used for cockroaches. The specific process is as follows:

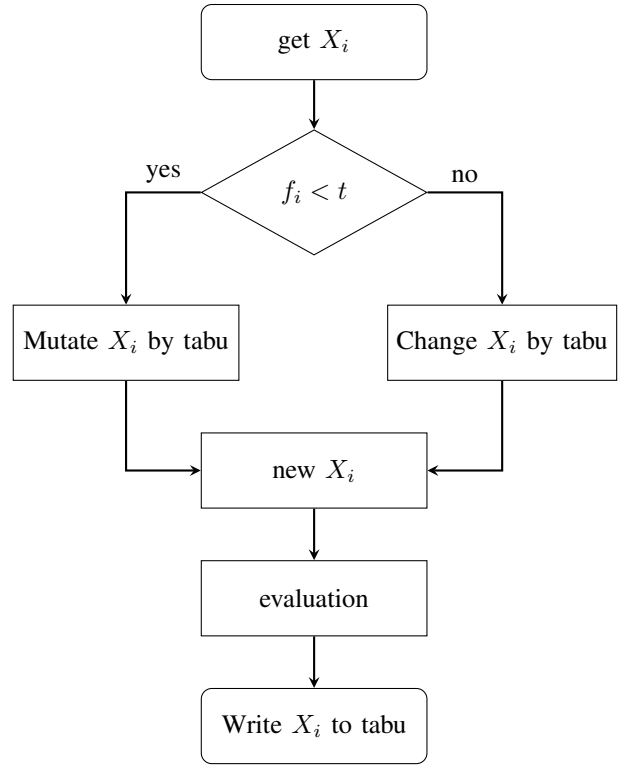
\begin{figure}
	\tikzstyle{startstop} = [rectangle, rounded corners, minimum width=3cm, minimum height=1cm,text centered, draw=black]
	\tikzstyle{decision} = [draw,diamond,minimum  width=3cm]
	\tikzstyle{process} = [rectangle, minimum width=3cm, minimum height=1cm, text centered, draw=black]
	\tikzstyle{arrow} = [thick,->,>=stealth]
	
	\begin{tikzpicture}[node distance = 1cm]
		\node (start) [startstop] {get $X_i$};

		\node (judge) [decision, below of = start,  yshift = -1cm] {$f_i < t$ };
		
		\node (mutation) [process, left of=judge, xshift=-1.5cm, yshift=-2cm] {Mutate $X_i$ by tabu};
		
		\node (gradient) [process, right of=judge, xshift=1.5cm, yshift=-2cm] {Change $X_i$ by tabu};
		
		\node (new) [process,  below of = judge,  yshift=-2.7cm, align=center] {new $X_i$};
		
		\node (evaluate) [process, below of=new, yshift=-0.7cm, align=center] {evaluation};
		
		\node (write2tabu) [startstop, below of=evaluate, yshift=-0.7cm, align=center] {Write $X_i$ to tabu};
		
		\draw [arrow] (start) -- (judge);
		\draw [arrow] (judge) -| node[near start, above] {yes} (mutation);
		\draw [arrow] (judge) -| node[near start, above] {no} (gradient);
		
		\draw [arrow](mutation) |- node[below]{}(new);
		\draw [arrow](gradient) |- node[below]{}(new);
		
		\draw [arrow](new) -- (evaluate);
		
		\draw [arrow](evaluate) -- (write2tabu);
		
	\end{tikzpicture}
	\caption{Flow chart of tabu search}
	\label{fig: Flow chart of fmincon AS algo}
	
\end{figure}

As can be seen from the flow chart, the tabu table is generated first and the initial solution is saved. Then, in the iterative process, the solutions are mutated or gradually changed according to the fitness value and tabu table, new solutions are generated for evaluation, and then the tabu table is written to continue the algorithm execution. The above measures can prevent the cockroach from accessing the same local optimal solution for a second time, thus avoiding the time spent by the algorithm to repeat the search.

\subsection{FEATURE SUBSETS AND POPULATION EVALUATION}

Subset evaluation criteria can be roughly divided into independent criteria and related criteria. In this paper, the prediction rates of kNN and SVM classifiers in related standards were used for evaluation. The classification model and feature selection depend on each other, and the accuracy of the feature subset is good. For each individual in the population, the fitness value is the prediction rate of the classifier, and the fitness value is used to evaluate the quality of the cockroach response. If the fitness value of cockroach i is greater than the previous fitness value of cockroach i during the search process, the individual optimal solution, individual fitness and the current position of cockroach (solution) are updated.

\subsection{IMIICSO ALGORITHM}

\begin{algorithm}[H]
	\caption{Feature selection algorithm based on cockroach swarm optimization and rough set.}\label{alg:alg1}
	\begin{algorithmic}[1]
		\REQUIRE Decision table $DT = (U,C \cup D )$
		\ENSURE Best feature subset $x_{best}$
		
		\STATE Random initial location of $X$ 
		\FOR{1 to N}
		\STATE Evaluate $X_i$
		\ENDFOR
		\STATE Generated inverse matrix $X^*$ 
		
		\STATE Choose N best solution

		\WHILE{ $iter \leq maxIter$ }
		
		\FOR{i = 1:N}
			
			\STATE Use tabu to mutate or change $X_i$
		
			\STATE Evaluate $X_i$ using classifiers
		
			\STATE Save $X_i$ to tabu
			\IF{accuracy >= fitness{j}}
				\STATE Update fitness(j)  and pbest
			\ELSE
				\STATE Use formula (10) to generate a new solution
			\ENDIF

		\ENDFOR

		\IF{  mod(iters,i) == 0 }
			\STATE Generate new solution based gbest
			\STATE Evaluate  using classifiers
		\ENDIF
		
		\STATE Update the optimal fitness and the optimal feature subset
		
		\STATE perform Dispersing and Ruthless behavior
		
		\ENDWHILE
		
		\RETURN $x_{best}$
	\end{algorithmic}
	\label{alg1}
\end{algorithm}

Algorithm 2 gives the pseudo-code of IMIICSO algorithm. Some details are given here.

1) As mentioned above, the classifier kNN(k=7) is used to evaluate all the solutions in the third row of algorithm 2 and update the applicability and global optimal solutions; In line 6, N optimal solutions are selected according to fitness to form the initial population.

2) In line 10, according to Formula (12), a new solution set is generated based on pbest and gbest neighborhood search;

3)11 Use tabu search. According to the suitability of the solution, the solution is mutated or gradually changed, and then regenerated into a new solution;

4) Line 24 achieves dispersion and cruelty through equations (7) and (8).

We analyze the time complexity of the proposed IMIICSO algorithm model. It can be seen from algorithm 1 that the time complexity of the algorithm depends on the cardinality size of conditional feature set C, while the time complexity of algorithm 2 has three main determinants: the number of iterations G, population size N, and the computation time of the classifier k. In the experiment, the time complexity of algorithm 1 depends largely on the number of instances. When the sample size is large, the time complexity of mutual information computation is relatively high. Therefore, the time complexity of IMIICSO algorithm is O(C + G*N*K).

\section{EXPERIMENTAL RESULTS AND ANALYSIS}

In this section, we will cover some of the details of the model implementation and present the experimental results and analysis. First, six data sets used in the experiment are introduced. Then the data is preprocessed; Then, the proposed feature selection algorithm based on cockroach swarm optimization and rough set is compared with the experimental results of the other three most representative algorithms in swarm optimization. Finally, the experimental results are discussed. All numerical experiments were carried out in MATLAB R2022b using MacBook Air, Apple M1 chip, MacOS 13.2beta, operating frequency of 3.20GHz, 8GB memory, and arm64 architecture. All algorithms use KNN classifier as evaluation criteria, and k=7 is determined on the basis of preliminary experimental classification.

\subsection{Experimental results and analysis}

Ten data sets downloaded from the UCI were used in the experiment, and basic statistics of these data sets are provided in the table.

\begin{table}
	\caption{Data description}
	\label{table}
	\setlength{\tabcolsep}{4pt}
	\begin{tabular}{p{33pt}p{33pt}p{33pt}p{33pt}p{33pt}p{33pt}}
		\hline   
		$No.$& $Datasets$& $Data type$& $|U|$& $|C|$& $|D|$ \\
		\hline
		$ 1 $& mushroom& Categorial & 8124 & 22 & 1\\
		$ 2 $& robot & Num & 5425 & 24  & 1\\
		$ 3 $& breast cancer & Num & 699 & 9  & 1\\
		$ 4 $& internet ads & Num & 3579& 1558 & 1 \\
		$ 5 $& clean & Num & 6598 & 166 & 1\\
		$ 6 $& audiology & Categorial & 200 & 68  & 1\\
		$ 7 $& darwin & Num & 174 & 451  & 1\\
		$ 8 $& isolet & Num & 6238 & 618 & 1 \\
		$ 9 $& Chess-king  & Categorial & 3196 & 37 & 1\\
		$ 10 $& spect & Num & 187 & 45  & 1\\
		\hline
	\end{tabular}
\end{table}

For each data set, sample set, conditional attribute set and decision attribute are expressed as U, C and D respectively. Part of the data set attribute values are text type and continuous variable type, which are normalized to discrete variable type. For null value data, mode replacement of column vector is adopted. In the algorithm, data set U(C; D), $70\%$is selected as the training sample, and $30\%$is left as the test sample, and the data set is randomly scrambled in each iteration to apply the cross-validation method. Group optimization algorithms ABC, ACO, PSO and our proposed IMIICSO algorithm are implemented on the decision table. The classifier KNN(k=7) in wrapper technology was used as the evaluation standard to calculate the classification accuracy of feature subset implementation.

\subsection{Parameter setting}

There are some important parameter Settings in the algorithm, but there is no theoretical proof of parameter setting rules at present. In this paper, all parameters are set according to experience.

The mutual information filtering threshold is 1:$m_i^1$, and the setting of this parameter has a great impact on the running time and accuracy of the algorithm model. The mutual information size distributions of conditional features and decision features in each data set are different, some distributions are concentrated, and some maximum mutual information and minimum mutual trust information differ greatly. $m_i^1=0.01$is used in the experiment.

The mutual information filtering threshold is 2:$m_i^2$. The feature corresponding to the mutual information below the threshold is filtered out. Both threshold value 1 and threshold value 2 have great influence on model running time.

Select the feature threshold: $f_t$, which is set as 0.5 according to the experiment. When the constructed solution vector is greater than or equal to 0.5, select the feature, and when it is less than, do not select.

\begin{table}
	\caption{IMIICSO parameter setting}
	\label{table}
	\setlength{\tabcolsep}{3pt}
	\begin{tabular}{p{115pt}p{115pt}}
		\hline   
		$Parameter information$ & $Setting rules$ \\
		\hline
		Mutual information threshold $m_i^1$ & 0~0.15, and step is 0.05 \\
		Mutual information threshold $m_i^2 $& $0~0.6$, change with $mi$ size \\
		Population number in CSO & Pop = 20 \\
		Total number of iterations & MAXITER = 100 \\
		Feature selection threshold & 0.5 \\
		\hline
	\end{tabular}
\end{table}

\subsection{Comparison with other feature selection models}

Feature selection is a very important step before data preprocessing tasks, and many scholars have studied different methods applied to it. In this paper, the cardinality, running time and accuracy of selected feature subsets are compared with several representative swarm intelligent optimization algorithms ABC swarm algorithm, ACO ant colony algorithm and PSO particle swarm optimization algorithm. The following is a brief description of these three population algorithms.

1)ABC[60] : a relatively new swarm intelligent optimization algorithm, which consists of honey source, hired bees and non-employed bees. Through the local optimization behavior of individual human worker bees, the optimal feature subset can be found by global optimization.

2)ACO[59] : It is a probabilistic algorithm to search for optimal path. It has the characteristics of distribution computation, positive feedback of information and heuristic search. Through the positive feedback mechanism, the search process is constantly converging and finally approaches the optimal solution.

3)PSO[65] : Through the strategy of searching the surrounding area of the bird closest to things, the speed and position of the bird are constantly updated during the search process to achieve the optimal solution search.

As mentioned above, repeated experiments have been conducted on three aspects. Table 8 describes the average number of optimal feature subsets selected by the algorithm RSCSO in 10 data sets. In addition, the average operation and average classification accuracy of the optimal feature subset obtained by the algorithm are described in Table 6 and 7 respectively. It should be pointed out that RSCSO running time includes rough intensive reduction time and random selection feature time for cockroach swarm optimization.

\begin{table}
	\caption{Comparison of classification accuracy using classifier kNN on 10 data sets}
	\label{table}
	\setlength{\tabcolsep}{4pt}
	\begin{tabular}{p{33pt}p{33pt}p{33pt}p{33pt}p{33pt}p{33pt}}
		\hline   
		Datasets & Raw data & ABC & ACO & PSO & IMIICSO \\
		\hline
		mushroom & 31.9524& 99.63 & 98.77 & 98.81 & 99.59 \\
		 robot & 79.5968 & 93.22 & 88.76 & 92.12  & 89.92\\
		 breast-cancer & 99.0476 & 99.524 & 99.524 & 99.524  & 99.524 \\
		internet-ads & 96.9512 & 99.79 & 99.187 & 99.23 & 99.289 \\
		clean & 96.2121 & 99.747 & 99.59 & 99.59 & 99.293 \\
		audiology & 50.0 & 83.333 & 80.0 & 81.667  & 85.0 \\
		 darwin & 54.717 & 88.679 & 83.019 & 84.906  & 90.566 \\
		 isolet & 90.7051 & 92.62 & 92.09 & 92.57 & 82.532 \\
		chess-king & 51.4077  & 92.492 & 80.0 & 89.966 & 94.8 \\
		spect & 73.6842 & 73.68 & 73.684 & 73.684  & 76.546 \\
		avg & 72.4274 & 92.2715 & 89.4624 & 90.9067  & 91.706 \\
		\hline
	\end{tabular}
\end{table}

\begin{table}
	\caption{Run time comparisons on 10 data sets}
	\label{table}
	\setlength{\tabcolsep}{4pt}
	\begin{tabular}{p{41pt}p{41pt}p{41pt}p{41pt}p{41pt}}
		\hline   
		Datasets &  ABC & ACO & PSO & IMIICSO \\
		\hline
		mushroom &  95.69 & 146.76 & 130.173 & 87.10 \\
		robot & 59.79 & 103.42 & 51.512  & 57.86 \\
		breast-cancer & 15.587 & 22.806 & 13.208  & 12.57 \\
		internet-ads &  1333.771 & 2365.684 & 834.735 & 334.673 \\
		clean & 483.062 & 961.292 & 238.487 & 273.265 \\
		audiology & 13.63 & 18.636 & 11.199  & 11.108 \\
		darwin & 11.249 &  22.072 & 11.323  & 12.144 \\
		isolet & 1274.0 & 4356.372 & 1047.856 & 424.675 \\
		chess-king & 73.987 & 32.04 & 35.697 & 39.809 \\
		spect & 15.154 & 20.176 & 9.712  & 19.308 \\
		\hline
	\end{tabular}
\end{table}

\begin{table}
	\caption{Cardinality comparison of selected feature subsets}
	\label{table}
	\setlength{\tabcolsep}{4pt}
	\begin{tabular}{p{41pt}p{41pt}p{41pt}p{41pt}p{41pt}}
		\hline   
		Datasets &  ABC & ACO & PSO & IMIICSO \\
		\hline
		mushroom &  8 & 7 & 8 & 5 \\
		robot & 4 & 7 & 6  & 4 \\
		breast-cancer & 6 & 6 & 5  & 3 \\
		internet-ads &  790 & 347 & 767 & 254 \\
		clean & 72 & 53 & 80 & 55 \\
		audiology & 34& 29 & 31  & 27 \\
		darwin & 242 &  315 & 226  & 179 \\
		isolet & 312 & 422 & 325 & 219 \\
		chess-king & 18 & 41 & 18 & 12 \\
		spect & 19 & 24 & 26 & 20 \\
		\hline
	\end{tabular}
\end{table}

\begin{table}
	\caption{Comparison of classification accuracy and feature subset cardinality using classifier SVM on 6 data sets}
	\label{table}
	\setlength{\tabcolsep}{4pt}
	\begin{tabular}{p{33pt}p{33pt}p{33pt}p{33pt}p{33pt}p{33pt}}
		\hline   
		Datasets & Raw data & ABC & ACO & PSO & IMIICSO \\
		\hline
		mushroom & 31.788& 99.344 & 98.195 & 98.523 & 99.365 \\
		breast-cancer & 89.523 & 99.048 & 98.095 & 98.571  & 99.048 \\
		internet-ads & 100 & 100 & 100 & 100 & 100 \\
		clean & 100 & 100 & 100 & 100 & 100 \\
		chess-king & 32.221  & 96.142 & 83.212 & 95.62 & 93.952 \\
		spect & 73.6842 & 73.684 & 73.684 & 73.684  & 73.684 \\
		avg & 42.7216 & 56.821 & 55.3186 & 90.9067  & 91.706 \\
		\hline
	\end{tabular}
\end{table}

As can be seen from the above table, in the classifier SVM, four data sets reported errors, so there was no test data. Compared with the original data sets, the proposed IMIICSO feature selection model achieves better classification performance on all data sets except the isolet data set. In addition, compared with the original data, the classification accuracy of IMIICSO model in kNN classifier and SVM is improved by $19.27\%$and $13.88\%$, respectively. This proves that removing weak correlation or redundant features can improve the classification performance of information systems. Compared with other feature selection models, IMIICSO can obtain fewer feature subsets and shorter running time while ensuring better classification accuracy. It can be seen from the above table that the growth of the cardinal random feature dimension of the selected feature subset increases substantially, and the running time increases with the increase of the number of instances and the cost increases. For the data set internet ads, the operation time is the longest, and the selected feature subset base is large, the number of ABC and PSO algorithm selection results reach more than 700, and the time is much longer. The proposed algorithm IMIICSO, with good classification accuracy guaranteed, selects only one third of the feature subset cardinality of other algorithms and one sixth of the data set. Table 7 shows that IMIICSO spends less time on all high-dimensional data sets. Especially for data sets with a large number of characteristics. In addition, for the large sample data set clean, the average running time of IMIICSO is almost the same as that of PSO due to the calculation of mutual information between the condition attribute and the decision attribute. Moreover, it can be seen from Table 6 and Table 7 that the average cardinality and average accuracy of the best feature subset obtained by IMIICSO are relatively good. In the case of ultra-high-dimensional data set, the feature with high mutual information is selected from the incoming feature set by algorithm 2. In the case of better classification accuracy, the computation running time and performance consumption can be effectively reduced.

\section{CONCLUSION}

$\quad  $In order to overcome the problem of high cost and low accuracy of feature selection algorithm in large scale and high dimensional data sets, we propose an IMIICSO method, which provides an effective method for feature selection based on mutual information or swarm intelligence. This method not only simplifies and speeds up the calculation of feature importance, but also reduces the cardinality of selected feature subset, thus improving the accuracy of selected feature subset and speeding up the speed of feature selection. As a result, execution times are significantly reduced, especially in high-dimensional data sets. The proposed method is compared with the existing group algorithm. The effectiveness and accuracy of the proposed method were verified by an experimental study on 10 data sets from UCI.

At present, IMIICSO method is easy to fall into local optimization for some data sets with small mutual information variance, and there will be redundant samples in large-scale data sets, which reduces accuracy and calculation speed. Therefore, future research will focus on designing data processing algorithms that consider both features and samples. In addition, we will try to combine fuzzy rough sets and fuzzy set theory to improve and propose a more efficient feature selection algorithm.

\section*{APPENDIX}
Appendixes, if needed, appear before the acknowledgment.

\section*{ACKNOWLEDGMENT}
The preferred spelling of the word ``acknowledgment'' in
American English is without an ``e'' after the ``g.'' Use the
singular heading even if you have many acknowledgments.
Avoid expressions such as ``One of us (S.B.A.) would like
to thank . . . .'' Instead, write ``F. A. Author thanks . . . .'' In
most cases, sponsor and financial support acknowledgments
are placed in the unnumbered footnote on the first page, not
here.


\section*{REFERENCES}

\begin{thebibliography}{17}

\bibitem{ref1} Pawlak.Z. Rough Sets[J].Int J Computer and Science, 1982, 11(5): 341-356.
\bibitem{ref2} Pawlak.Z. Rough Sets Theory and It's Applications to Date Analysis[J]. Cybernetnetics and Systems, 1998, 29(4): 661-688. 
\bibitem{ref5} Wang guo yin,Yu hong,Yang Da Chun.Decision table reduction based on conditional information entroy[J]. Chinese journal of Computers, 2002, 25(7): 759-766. 
\bibitem{ref10}Estévez P A, Tesmer M, Perez C A, et al. Normalized mutual information feature selection[J]. IEEE Transactions on neural networks, 2009, 20(2): 189-201.
\bibitem{ref13} Yang M,Chen S C,Yang X B. A novel approach of rough set-based attribute reduction using fuzzy discernibility matrix[C]. FSKD'07.Haikou, 2007, 3: 96-101. 
\bibitem{ref14} Z. Xu, C. Zhang, S. Zhang, W. Song, and B. Yang, ‘‘Effificient attribute reduction based on discernibility matrix,’’ in Proc. Int. Conf. Rough Sets Knowl. Technol. Berlin, Germany: Springer, May 2007, pp. 13–21. 
\bibitem{ref15} D. Chen, S. Zhao, L. Zhang, Y. Yang, and X. Zhang, ‘‘Sample pair selection for attribute reduction with rough set,’’ IEEE Trans. Knowl. Data Eng., vol. 24, no. 11, pp. 2080–2093, Nov. 2012. 
\bibitem{ref16} C. Wang, Q. He, M. Shao, and Q. Hu, ‘‘Feature selection based on maximal neighborhood discernibility,’’ Int. J. Mach. Learn. Cybern., vol. 9, no. 11, pp. 1929–1940, Nov. 2018. 
\bibitem{ref17}Theresa M. Korn; Korn, Granino Arthur. Mathematical Handbook for Scientists and Engineers: Definitions, Theorems, and Formulas for Reference and Review. New York: Dover Publications. ISBN 0-486-41147-8.
\bibitem{ref18} Yeung D S, Chen D, Tsang E C C, et al. On the generalization of fuzzy rough sets[J]. IEEE Transactions on fuzzy systems, 2005, 13(3): 343-361.
\bibitem{ref19} Haoyang, W., et al. (2009). "An Interval Type-2 Fuzzy Rough Set Model for Attribute Reduction." IEEE Transactions on Fuzzy Systems 17(2): 301-315.
\bibitem{ref20} Wu H, Wu Y, Luo J. An interval type-2 fuzzy rough set model for attribute reduction[J]. IEEE Transactions on fuzzy systems, 2009, 17(2): 301-315.
\bibitem{ref21} Atanassov K T, Atanassov K T. Interval valued intuitionistic fuzzy sets[J]. Intuitionistic Fuzzy Sets: Theory and Applications, 1999: 139-177.
\bibitem{ref22} Brown J G. A note on fuzzy sets[J]. Information and control, 1971, 18(1): 32-39.
\bibitem{ref23} Nowicki R. On combining neuro-fuzzy architectures with the rough set theory to solve classification problems with incomplete data[J]. IEEE Transactions on Knowledge and Data Engineering, 2008, 20(9): 1239-1253.
\bibitem{ref24} Zadeh L A. Fuzzy sets[J]. Information and control, 1965, 8(3): 338-353.
\bibitem{ref25} Hu Q, Zhang L, An S, et al. On robust fuzzy rough set models[J]. IEEE transactions on Fuzzy Systems, 2011, 20(4): 636-651.
\bibitem{ref26}Jensen R, Shen Q. New approaches to fuzzy-rough feature selection[J]. IEEE Transactions on fuzzy systems, 2008, 17(4): 824-838.
\bibitem{ref27} Zhang X, Mei C, Chen D, et al. Active incremental feature selection using a fuzzy-rough-set-based information entropy[J]. IEEE Transactions on Fuzzy Systems, 	2019, 28(5): 901-915.
\bibitem{ref28} Sun L, Wang L, Ding W, et al. Feature selection using fuzzy neighborhood entropy-based uncertainty measures for fuzzy neighborhood multigranulation rough sets[J]. IEEE Transactions on Fuzzy Systems, 2020, 29(1): 19-33.
\bibitem{ref29} X. Zhang, C. Mei, D. Chen, and J. Li, ‘‘Feature selection in mixed data: A method using a novel fuzzy rough set-based information entropy,’’ Pattern Recognit., vol. 56, pp. 1–15, Aug. 2016. 
\bibitem{ref30} J. Liang and Z. Z. Shi, ‘‘The information entropy, rough entropy and knowledge granulation in rough set theory,’’ Int. J. Uncertainty, Fuzziness Knowl.-Based Syst., vol. 12, no. 1, pp. 37–46, 2004. 
\bibitem{ref31} C. Gao, Z. Lai, J. Zhou, C. Zhao, and D. Miao, ‘‘Maximum decision entropy-based attribute reduction in decision-theoretic rough set model,’’ Knowl.-Based Syst., vol. 143, pp. 179–191, Mar. 2018. 
\bibitem{ref32} X. Zhang, C. Mei, D. Chen, Y. Yang, and J. Li, ‘‘Active incremental feature selection using a fuzzy-rough-set-based information entropy,’’ IEEE Trans. Fuzzy Syst., vol. 28, no. 5, pp. 901–915, May 2020. 
\bibitem{ref33} Zhao H, Wang P, Hu Q, et al. Fuzzy rough set based feature selection for large-scale hierarchical classification[J]. IEEE Transactions on Fuzzy Systems, 2019, 	27(10): 1891-1903.
\bibitem{ref34} Kong L, Qu W, Yu J, et al. Distributed feature selection for big data using fuzzy 	rough sets[J]. IEEE Transactions on Fuzzy Systems, 2019, 28(5): 846-857.
\bibitem{ref35} Wang C, Hu Q, Wang X, et al. Feature selection based on neighborhood discrimination index[J]. IEEE transactions on neural networks and learning systems, 2017, 29(7): 2986-2999.
\bibitem{ref36} Wang C, Huang Y, Shao M, et al. Feature selection based on neighborhood self-information[J]. IEEE Transactions on Cybernetics, 2019, 50(9): 4031-4042.
\bibitem{ref37} Yang Y, Chen D, Wang H, et al. Incremental perspective for feature selection based on fuzzy rough sets[J]. IEEE Transactions on Fuzzy Systems, 2017, 26(3): 	1257-1273.
\bibitem{ref38} Jahani M S, Aghamollaei G, Eftekhari M, et al. Unsupervised feature selection 	guided by orthogonal representation of feature space[J]. Neurocomputing, 2023, 	516: 61-76.
\bibitem{ref39} Yang Y, Chen D, Zhang X, et al. Incremental feature selection by sample selection 	and feature-based accelerator[J]. Applied Soft Computing, 2022, 121: 108800.
\bibitem{ref40} Jahani M S, Aghamollaei G, Eftekhari M, et al. Unsupervised feature selection 	guided by orthogonal (Nowicki 2008)representation of feature space[J]. Neurocomputing, 2023, 	516: 61-76.
\bibitem{ref41} Xie W, Wang L, Yu K, et al. Improved multi-layer binary firefly algorithm for 	optimizing feature selection and classification of microarray data[J]. Biomedical 	Signal Processing and Control, 2023, 79: 104080.
\bibitem{ref42} Chen D, Yang Y. Attribute reduction for heterogeneous data based on the 	combination of classical and fuzzy rough set models[J]. IEEE Transactions on Fuzzy Systems, 2013, 22(5): 1325-1334.
\bibitem{ref43} Tan A, Wu W Z, Qian Y, et al. Intuitionistic fuzzy rough set-based granular 	structures and attribute subset selection[J]. IEEE Transactions on Fuzzy Systems, 2018, 27(3): 527-539.
\bibitem{ref44} Zhao S, Tsang E C C, Chen D. The model of fuzzy variable precision rough sets[J]. 	IEEE transactions on Fuzzy Systems, 2009, 17(2): 451-467.
\bibitem{ref45} Mi J S,Wu W Z,Zhang W X. Approaches to knowledge reduction based on variable precision rough sets model.Information Sciences[J], 2004, 255-272.
\bibitem{ref46} Ziarko W.Variable precision rough sets model.Journal of computer and systems Sciences[J], 1993, 46(1): 39-59. 
\bibitem{ref47} Greco S,Matarazzo B,Slowingski R. A new rough set approach to multicriteria and multiattribute classification.Rough sets and current trends in computing[M], 1424, 1998, 60-67.
\bibitem{ref48} Wang C, Wang Y, Shao M, et al. Fuzzy rough attribute reduction for categorical data[J]. IEEE Transactions on Fuzzy Systems, 2019, 28(5): 818-830.
\bibitem{ref49} TsangE C C,Chen D,Yeung D S. Attributes reduction using fuzzy rough sets[J]. IEEE Transaction on Fuzzy Systems, 2005, 13(3): 343-361. 
\bibitem{ref50} Bhuvaneshwari M, Kanaga E G M, Anitha J. Bio-inspired Red Fox-Sine cosine 	optimization for the feature selection of SSVEP-based EEG signals for BCI applications[J]. Biomedical Signal Processing and Control, (De Cock, Cornelis et al. 2007)
\bibitem{ref51} Eskandari S, Seifaddini M. Online and offline streaming feature selection methods with bat algorithm for redundancy analysis[J]. Pattern Recognition, 2023, 133:109007.
\bibitem{ref52} Nowicki, R. (2008). "On Combining Neuro-Fuzzy Architectures with the Rough Set Theory to Solve Classification Problems with Incomplete Data." IEEE Transactions on Knowledge and Data Engineering 20(9): 1239-1253.
\bibitem{ref53} De Cock, M., et al. (2007). "Fuzzy Rough Sets: The Forgotten Step." IEEE Transactions on Fuzzy Systems 15(1): 121-130.
\bibitem{ref54} Wang Xi Zhao,Ha Yan,Chen De Gang. On the reduction of fuzzy rough sets.In:Proceeding of the Third International Conference on Machine Learning and Cybernetics[C], Guangzhou, 2005, 18-21: 3175-3178. 
\bibitem{ref55} Xiaodong, L., et al. (2009). "The Development of Fuzzy Rough Sets with the Use of Structures and Algebras of Axiomatic Fuzzy Sets." IEEE Transactions on Knowledge and Data Engineering 21(3): 443-462.
\bibitem{ref59} Dorigo M. Optimization, learning and natural algorithms[J]. Ph. D. Thesis, Politecnico di Milano, 1992.
\bibitem{ref60} Karaboga D. Artificial bee colony algorithm[J]. scholarpedia, 2010, 5(3): 6915.
\bibitem{ref62} Tiwari, S. and S. Agarwal (2023). "Empirical analysis of chronic disease dataset for multiclass classification using optimal feature selection based hybrid model with spark streaming." Future Generation Computer Systems 139: 87-99.
\bibitem{ref63}Thakkar, A. and R. Lohiya (2023). "Fusion of statistical importance for feature selection in Deep Neural Network-based Intrusion Detection System." Information Fusion 90: 353-363.
\bibitem{ref64} Zhang, Y., et al. (2023). "Recognising drivers’ mental fatigue based on EEG multi-dimensional feature selection and fusion." Biomedical Signal Processing and Control 79.
\bibitem{ref65} Kennedy J, Eberhart R. Particle swarm optimization[C]//Proceedings of ICNN'95-international conference on neural networks. IEEE, 1995, 4: 1942-1948.
\bibitem{ref66} Song X F, Zhang Y, Gong D W, et al. A fast hybrid feature selection based on correlation-guided clustering and particle swarm optimization for high-dimensional data[J]. IEEE Transactions on Cybernetics, 2021, 52(9): 9573-9586.
\bibitem{ref67} Greenbank, S. and D. A. Howey (2023). "Piecewise-linear modelling with automated feature selection for Li-ion battery end-of-life prognosis." Mechanical Systems and Signal Processing 184.
\bibitem{ref68} Ewees, A. A., et al. (2023). "Gradient-based optimizer improved by Slime Mould Algorithm for global optimization and feature selection for diverse computation problems." Expert Systems with Applications 213.
\bibitem{ref69} Gégény, D. and S. Radeleczki (2022). "Rough L-fuzzy sets: Their representation and related structures." International Journal of Approximate Reasoning 142: 1-12.
\bibitem{ref70} Cheng L, Han L, Zeng X, et al. Adaptive cockroach colony optimization for rod-like robot navigation[J]. Journal of Bionic Engineering, 2015, 12(2): 324-337.
\bibitem{ref71} Wu, S. J. and C. T. Wu (2013). "Computational optimization for S-type biological systems: cockroach genetic algorithm." Math Biosci 245(2): 299-313.
\bibitem{ref72} Hu, Q., et al. (2011). "Kernelized Fuzzy Rough Sets and Their Applications." IEEE Transactions on Knowledge and Data Engineering 23(11): 1649-1667.
\bibitem{ref73} Huang, Z., et al. (2022). "Noise-Tolerant Fuzzy-$\beta$-Covering-Based Multigranulation Rough Sets and Feature Subset Selection." IEEE Transactions on Fuzzy Systems 30(7): 2721-2735.




\end{thebibliography}
\end{document}